\documentclass[letterpaper,10pt,conference]{ieeeconf}

\IEEEoverridecommandlockouts
\pdftrailerid{}
\usepackage{amsmath}
\usepackage{amssymb}
\usepackage{booktabs}
\usepackage[nocompress]{cite}
\usepackage{graphicx}
\usepackage{url}
\graphicspath{{figures/}}

\title{\LARGE \bf
AeroDASH: Asynchronous Vision-Language-Action Control for Continuous UAV Flight
}

\author{Zhe Liu$^{1,2}$, Yuxuan Zheng$^{2}$, Jinzhe Qin$^{2}$, Ziqi Gao$^{1}$, and Boyu Zhou$^{2,3,\dagger}$\thanks{$^{\dagger}$Corresponding Author.}\thanks{\raggedright $^{1}$Shenzhen International Graduate School, Tsinghua University, Shenzhen 518055, China (e-mail: \mbox{z-liu26@mails.tsinghua.edu.cn}; \mbox{ziqigao@sz.tsinghua.edu.cn}).}\thanks{\raggedright $^{2}$Department of Mechanical and Energy Engineering, Southern University of Science and Technology, Shenzhen, China (e-mail: \mbox{12511006@mail.sustech.edu.cn}; \mbox{12412111@mail.sustech.edu.cn}; \mbox{zhouby@sustech.edu.cn}).}\thanks{$^{3}$Differential Robotics.}}

\begin{document}

\maketitle
\thispagestyle{empty}
\pagestyle{empty}

\begin{abstract}
Onboard inference latency creates a mismatch between the trajectories predicted
by vision--language--action (VLA) policies and the ongoing motion of unmanned
aerial vehicles (UAVs).
We present AeroDASH, an asynchronous VLA framework for continuous
language-conditioned flight. A B-spline policy is initialized through supervised
fine-tuning on expert demonstrations, providing continuous trajectory references
in position and yaw. During execution, trajectory fusion removes the expired
prediction prefix and connects the remaining segment to the active reference
with derivative continuity while limiting trajectory deformation.
To account for deviations induced by asynchronous execution,
execution-aware group relative policy optimization (GRPO) further adapts the policy
using trajectory-level rewards. On the UAV-Flow test set, AeroDASH achieves a
success rate of 76.56\%, exceeding OpenVLA-UAV and AeroDASH w/o GRPO by 38.1 and
20.1 percentage points, respectively. Execution experiments show that AeroDASH
sustains motion during inference and reduces acceleration, jerk, and speed
fluctuations relative to asynchronous tracking of waypoint sequences.
We further demonstrate language-conditioned flight with onboard inference and
asynchronous trajectory execution. The code will be released upon acceptance.
\end{abstract}

\section{Introduction}
\label{sec:introduction}

\suppressfloats[t]
\begin{figure}[t]
  \centering
  \includegraphics[width=\columnwidth]{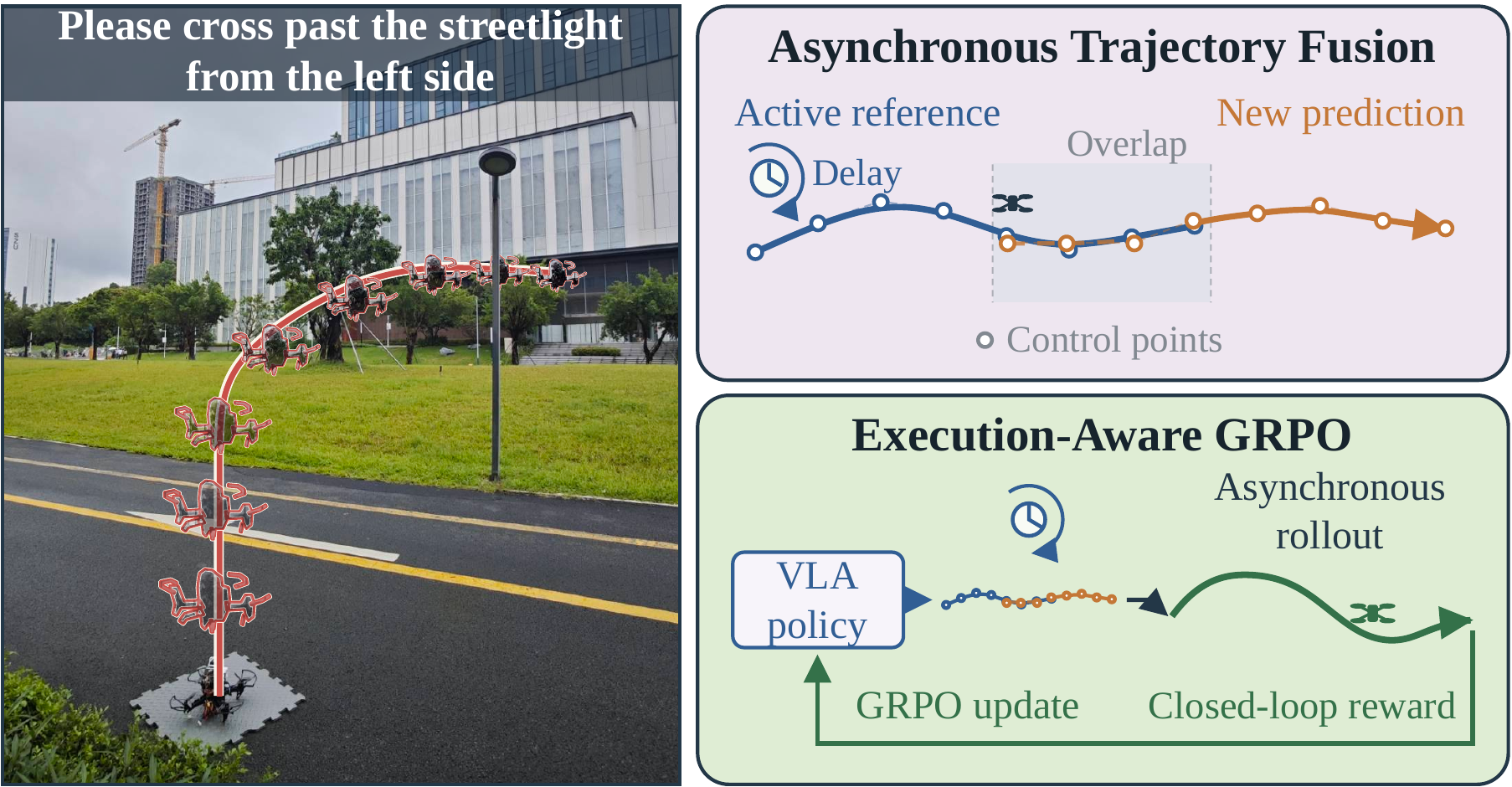}
  \caption{AeroDASH. The left panel illustrates language-conditioned flight in a real-world scene.
  Asynchronous trajectory fusion connects delayed B-spline predictions to the
  active reference. Execution-aware GRPO updates the policy using rewards from
  realized asynchronous flights. Hollow markers denote B-spline control points.}
  \label{fig:aerodash-overview}
\end{figure}

Language-conditioned flight requires an unmanned aerial vehicle (UAV) to ground natural language instructions in visual observations and translate them into continuous three-dimensional motion. Vision--language--action (VLA) models have emerged as a promising paradigm for this task, learning to map visual observations and language commands directly to robot actions. However, continuous UAV flight imposes a stringent execution requirement that goes beyond discrete action prediction: the flight controller needs a temporally smooth reference between successive policy updates, with high-order derivative continuity across all degrees of freedom in the vehicle's flat-output space.

Current VLA approaches for UAVs predict discrete waypoints or pose targets, leaving the system without a continuous reference between updates. Onboard inference latency further exacerbates this gap: during VLA inference, the UAV continues to move, so each delayed prediction is conditioned on a stale observation and arrives after part of its intended horizon has elapsed. Existing asynchronous execution strategies address this mismatch at the position level only, without enforcing the high-order derivative continuity that UAV flight controllers require across the flat-output space. Abrupt transitions in these derivatives produce large jerks that can destabilize the vehicle. A second gap lies in the mismatch between open-loop training and closed-loop execution. Supervised fine-tuning trains the policy to match individual predictions to expert demonstrations but does not directly evaluate the complete trajectory produced by asynchronous execution.

To overcome these challenges, we present AeroDASH (Fig.~\ref{fig:aerodash-overview}), an asynchronous VLA framework for UAVs that sustains continuous language-conditioned flight under onboard inference latency. To address the first gap, the VLA policy predicts control points that define a cubic B-spline over the quadrotor's four-dimensional flat-output space, providing an analytically differentiable reference throughout each inference interval. When a delayed proposal arrives, the portion preceding the scheduled switch time is removed and the remaining segment is refitted to enforce high-order derivative continuity with the active reference. The fusion formulation enforces $C^2$ continuity of the trajectory reference at switching boundaries while penalizing deviations from the retained prediction. To address the second gap, we introduce execution-aware post-training with group relative policy optimization (GRPO). We optimize the policy using trajectory-level rewards that measure agreement between the complete trajectories produced by asynchronous execution and expert demonstrations.

\begin{itemize}
  \item We design an asynchronous B-spline fusion mechanism that aligns delayed predictions with execution time and enforces $C^2$ continuity at switching boundaries while limiting trajectory deformation.
  \item We introduce GRPO post-training with trajectory-level rewards computed from asynchronously executed trajectories to improve agreement with expert demonstrations.
\end{itemize}
\section{Related Work}
\label{sec:related_work}

\subsection{Vision-Language-Action for Drones}
General-purpose VLA frameworks learn language-conditioned robot actions from
diverse embodied demonstrations~\cite{zitkovich2023rt2,kim2025openvla,
black2025pi0,bu2025univla}. For UAVs, existing work spans long-horizon
language-guided navigation and fine-grained flight control.
AerialVLN~\cite{liu2023aerialvln} studies navigation toward a distant goal,
whereas UAV-Flow Colosseo~\cite{wang2025uavflow} pairs individual flight
instructions with pilot demonstrations and trains VLA policies to predict
UAV poses. Its Flying-on-a-Word tasks require the flight path to follow the
spatial relationships specified by the instruction, making the executed path
relevant in addition to the final position.

Reinforcement learning extends VLA training beyond demonstration imitation.
SimpleVLA-RL~\cite{li2026simplevlarl} applies outcome-driven reinforcement
learning to OpenVLA-OFT, using online trajectory sampling and exploration to
improve manipulation performance and generalization. Its formulation provides
feedback on the consequences of sequential actions, rather than supervising
each prediction solely against demonstrated actions. In aerial navigation,
EG-GRPO~\cite{chen2026eggrpo} augments online rollouts with expert data to guide
exploration and improve alignment with fine-grained instructions.

These studies establish reinforcement learning as a means of improving
instruction-conditioned behavior, while AeroDASH focuses on the effect of
asynchronous execution on the resulting flight trajectory. The policy predicts
B-spline control points and receives GRPO updates from closed-loop trajectory
returns. This training objective evaluates successive predictions through
their execution outcomes, complementing the temporal alignment and trajectory
fusion discussed next.

\subsection{Asynchronous Trajectory Execution}
Optimization-based trajectory generation provides a foundation for continuous
UAV flight. Minimum-snap methods~\cite{mellinger2011minimumsnap} penalize
high-order derivatives of polynomial trajectories to generate smooth motion
through prescribed waypoints. Richter et al.~\cite{richter2016polynomial}
develop an efficient formulation for jointly optimizing polynomial segments
and allocating their durations. MINCO~\cite{wang2022minco} further supports
efficient trajectory optimization under geometric and dynamic constraints.
These methods optimize flight trajectories from planning objectives and
constraints; AeroDASH instead uses optimization to correct a trajectory
predicted by a language-conditioned policy while limiting changes to its shape.

Receding-horizon planners update trajectories as the flight state evolves.
B-spline-based systems exploit local support and analytic derivatives for
online kinodynamic replanning~\cite{usenko2017replanning,ding2019bspline,
zhou2019robust,zhou2021egoplanner}, while FASTER~\cite{tordesillas2022faster}
maintains a committed trajectory during successive planning cycles. These
mechanisms allow flight to continue during planning. For a VLA, however, a
prediction is conditioned on an earlier visual observation and may arrive
after part of its intended execution interval has elapsed. Maintaining an
executable trajectory during inference therefore does not by itself resolve
how the delayed prediction should be aligned and connected to it.

RTC and REMAC address asynchronous execution through action generation.
RTC~\cite{black2025rtc} guides new predictions using overlapping actions,
whereas REMAC~\cite{wang2026remac} learns delay-aware corrections and preserves
the executed prefix during sampling. Both promote consistency between
successive predictions, but neither explicitly imposes velocity and
acceleration matching at the switching boundary.

AeroDASH aligns delayed B-spline predictions with execution time and refits
the remaining curve to match the current reference's position, velocity,
and acceleration while limiting deformation. GRPO post-training further
adapts the policy to asynchronous execution using closed-loop trajectory returns.
\section{Method}
\label{sec:method}

\begin{figure*}[t]
  \centering
  \vspace*{5pt}
  \includegraphics[width=\textwidth]{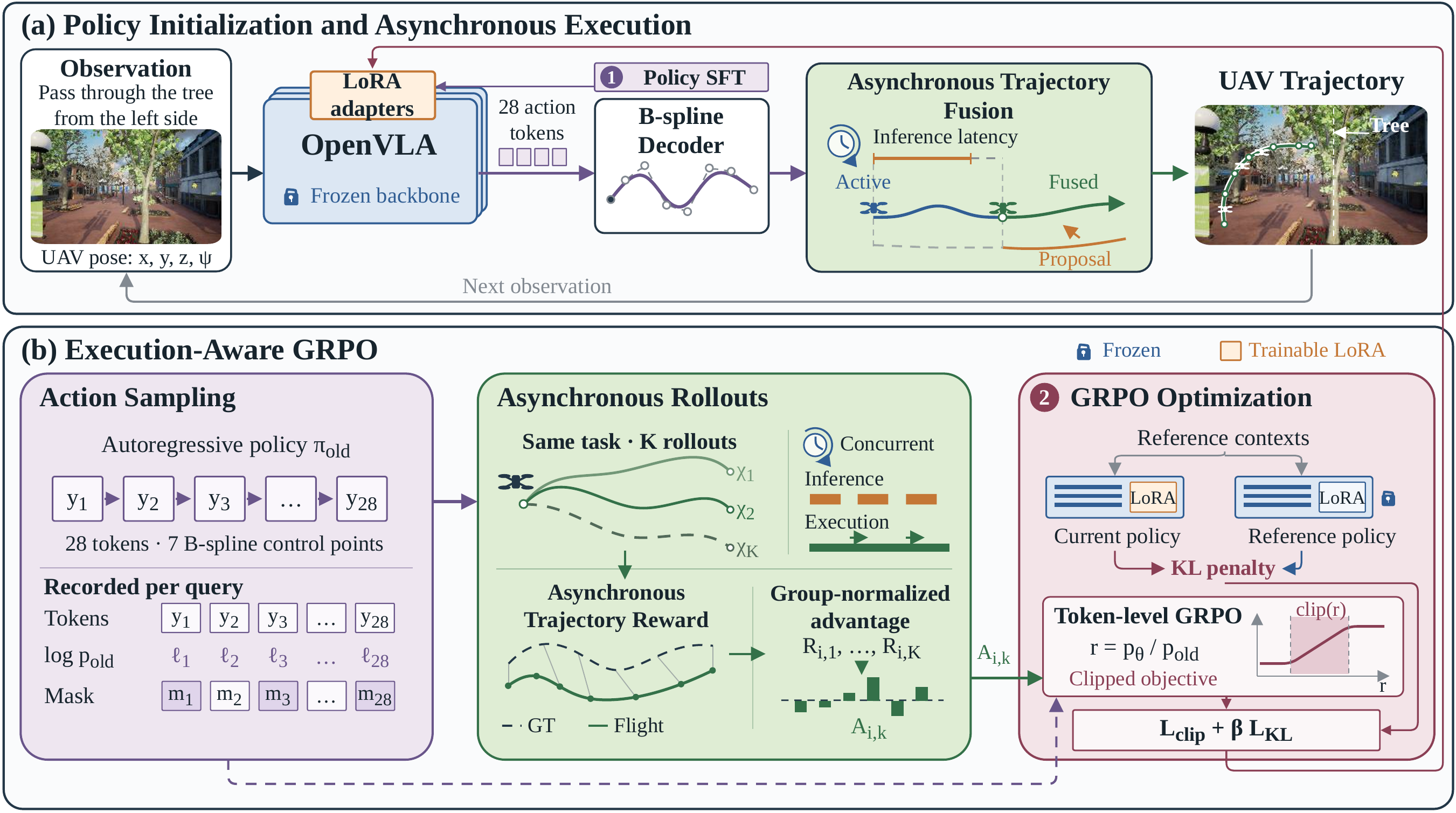}
\caption{Overview of AeroDASH.
(a) SFT initializes the policy using tokenized B-spline control points fitted
to expert flight trajectories. The onboard policy decodes B-spline proposals, which the asynchronous
executor aligns and fuses with the active reference.
(b) Closed-loop flight rewards yield group-normalized advantages, while
recorded action tokens and pre-update log probabilities support token-level clipped
GRPO. Policy updates modify only the LoRA adapters, with KL regularization
against a frozen reference policy.}
  \label{fig:aerodash-training}
\end{figure*}

Figure~\ref{fig:aerodash-training} outlines AeroDASH's training and execution
pipeline. Supervised fine-tuning (SFT) initializes the B-spline policy from expert trajectories.
Because the UAV continues moving during inference, optimization-based fusion
aligns delayed proposals with execution time and connects them continuously
to the active reference. GRPO then adapts the policy to accumulated closed-loop
deviations using trajectory-level rewards.

\subsection{Preliminaries}
\label{sec:preliminaries}
AeroDASH uses an autoregressive OpenVLA policy~\cite{kim2025openvla}
to generate B-spline trajectory proposals. At query time $t_q$, the policy receives
$o_q=(I_q,\ell,\mathbf{s}_q)$, where $I_q$ is a forward-facing RGB image,
$\ell$ is the flight instruction, and
$\mathbf{s}_q=[x_q,y_q,z_q,\psi_q]^\top$ contains the UAV position and yaw.
Position and yaw form a set of flat outputs for the quadrotor
model~\cite{faessler2018flatness}. We parameterize trajectories in this
four-dimensional output space.
The policy generates 28 action tokens
$\mathbf{y}_q=(y_{q,1},\ldots,y_{q,28})$ according to
\begin{equation}
    \pi_\theta(\mathbf{y}_q\mid o_q)
    =\prod_{j=1}^{28}
      p_\theta\!\left(y_{q,j}\mid o_q,\mathbf{y}_{q,<j}\right),
    \label{eq:autoregressive_policy}
\end{equation}
where $\theta$ denotes the policy parameters and $\mathbf{y}_{q,<j}$ contains
the preceding action tokens. Detokenization and denormalization map these
tokens to seven four-dimensional B-spline control points
$\mathbf{C}_{q,i}^{b}=[x_{q,i}^{b},y_{q,i}^{b},z_{q,i}^{b},
\psi_{q,i}^{b}]^\top\in\mathbb{R}^{4}$, $i=1,\ldots,7$.
Superscript $b$ denotes a local frame centered at the UAV position and aligned
with its heading at $t_q$, with the vertical axis unchanged. Adding the fixed
control point $\mathbf{C}_{q,0}^{b}=\mathbf{0}$ yields eight controls that
jointly parameterize a continuous flat-output trajectory.

Let $\mathbf{C}_{q,i}$ denote the corresponding world-frame controls, obtained
using the coordinate transformation in Sec.~\ref{sec:trajectory-generation}.
These controls define a clamped cubic B-spline with uniformly spaced interior
knots~\cite{ding2019bspline,zhou2021egoplanner}:
\begin{equation}
    \hat{\boldsymbol{\tau}}_q(u)
    =\sum_{i=0}^{7}N_{i,3}(u)\mathbf{C}_{q,i},
    \qquad u\in[0,1],
    \label{eq:bspline_proposal}
\end{equation}
where $N_{i,3}$ denotes the cubic B-spline basis function. The normalized phase
$u$ corresponds to physical time $t=t_q+uT_H$, where $T_H>0$ is the proposal
duration.

\subsection{Continuous Trajectory Generation}
\label{sec:trajectory-generation}
To learn continuous flight trajectories from demonstrations, the VLA needs
supervision in its control-point output representation. We fit B-splines to
expert flight segments and use the fitted control points as prediction targets.
We then train the policy through SFT using low-rank adaptation (LoRA).
Each demonstration segment contains $M=8$ flat-output samples, expressed in the
query-local frame defined in Sec.~\ref{sec:preliminaries} and arranged as
$\mathbf{S}_q^b\in\mathbb{R}^{M\times4}$. For spline fitting, we map the sample
indices to $[0,1]$ using $u_m=(m-1)/(M-1)$. Before fitting, yaw angles are
unwrapped to avoid artificial discontinuities at the angle wrap boundary. Let
$\bar{\boldsymbol{\Phi}}\in\mathbb{R}^{M\times7}$ have entries
$\bar{\Phi}_{m,i}=N_{i,3}(u_m)$, $i=1,\ldots,7$. With
$\mathbf{C}_{q,0}^b=\mathbf{0}$, the seven control-point targets are obtained
by least-squares fitting:
\begin{equation}
    \mathbf{C}_{q,1:7}^{b,*}
    =
    \arg\min_{\mathbf{C}\in\mathbb{R}^{7\times4}}
    \left\|
        \bar{\boldsymbol{\Phi}}\mathbf{C}
        -\mathbf{S}_q^{b}
    \right\|_{F}^{2}.
    \label{eq:control_target_fit}
\end{equation}
The rows of $\mathbf{C}_{q,1:7}^{b,*}$ correspond to the seven predicted
control points.

The fitted control points are normalized and tokenized to form the target sequence
$\mathbf{y}_q^{*}$. The action-token prediction loss is
\begin{equation}
    \mathcal{L}_{\mathrm{act}}
    =
    -\frac{1}{28}
    \sum_{j=1}^{28}
    \log p_{\theta}
    \left(
        y_{q,j}^{*}
        \mid
        o_q,\mathbf{y}_{q,<j}^{*}
    \right).
    \label{eq:action_sft}
\end{equation}

At inference, the decoded controls are transformed into the world frame using
the query position $\mathbf{p}_q=[x_q,y_q,z_q]^\top$ and yaw $\psi_q$:
\begin{equation}
\begin{aligned}
    \mathbf{p}_{q,i}^{w}
       &=\mathbf{R}_{z}(\psi_q)\mathbf{p}_{q,i}^{b}+\mathbf{p}_q,\\
    \psi_{q,i}^{w}
       &=\psi_{q,i}^{b}+\psi_q,
\end{aligned}
\qquad i=0,\ldots,7.
\label{eq:query_to_world}
\end{equation}
Here, $\mathbf{p}_{q,i}^b$ is the position component of the local control point,
and $\mathbf{R}_z(\psi_q)$ denotes rotation about the vertical axis. The
transformed controls
$\mathbf{C}_{q,i}=[(\mathbf{p}_{q,i}^w)^\top,\psi_{q,i}^w]^\top$ define the
world-frame proposal in~\eqref{eq:bspline_proposal}.

\subsection{Asynchronous Trajectory Fusion}
\label{sec:trajectory-fusion}

\begin{figure*}[t]
  \centering
  \includegraphics[width=\textwidth]{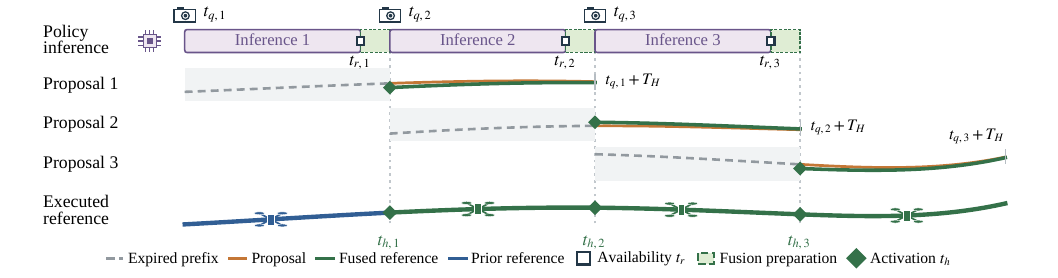}
  \caption{Asynchronous trajectory fusion over three successive predictions.
  For prediction $i$, inference starts at $t_{q,i}$ and the result becomes
  available at $t_{r,i}$,
  while the current reference continues to execute until the scheduled switch
  at $t_{h,i}$. The gray dashed prefix expires before the switch; the remaining
  prediction (orange) is corrected to form the fused reference (green),
  retaining its original end time $t_{q,i}+T_H$. The bottom row shows the
  resulting reference across switches. Green dashed boxes indicate preparation
  windows.}
  \label{fig:async-trajectory-timing}
\end{figure*}

SFT learns expert trajectories but does not enforce continuity between delayed
predictions and the reference being tracked. Since the UAV keeps moving during
inference, the returned prediction must first be aligned with the scheduled
switch time. Temporal alignment alone does not resolve differences in the flat
outputs and their first two physical-time derivatives at the switch.
Fusion therefore adjusts the remaining prediction to match this boundary,
avoiding abrupt changes in the reference sent to the controller.

A query at $t_q$ produces $\hat{\boldsymbol{\tau}}_q$ at $t_r$, after an
inference delay of $t_r-t_q$ (Fig.~\ref{fig:async-trajectory-timing}).
The UAV continues tracking the current reference
$\boldsymbol{\tau}^{\mathrm{act}}$ throughout inference and preparation,
switching to the corrected proposal at $t_h$ only if accepted.
The current reference must cover this entire interval.
\begin{equation}
\begin{aligned}
    t_q&\leq t_r\leq t_h<t_q+T_H,\\
    \Delta_q&=t_h-t_q, \qquad
    u_q=\frac{\Delta_q}{T_H}.
\end{aligned}
    \label{eq:async_clock}
\end{equation}
Here, $\Delta_q$ includes inference and any further waiting before activation;
$u_q$ is the elapsed fraction of the prediction horizon.
By $t_h$, the part of the prediction assigned to $[t_q,t_h)$ has expired.
We remove this part and, for $u_q<1$, keep the segment assigned to
$[t_h,t_q+T_H]$. In the prediction's original time coordinate $u$, the retained
segment occupies $[u_q,1]$.

To express the retained segment on the same $[0,1]$ interval used by the spline
representation, we introduce $v$, with $v=0$ at $t_h$ and $v=1$ at $t_q+T_H$:
\begin{equation}
\begin{aligned}
    \hat{\boldsymbol{\tau}}_q^{+}(v)
       &=\hat{\boldsymbol{\tau}}_q
          \!\left(u_q+(1-u_q)v\right), \quad v\in[0,1],\\
    T_q^{+}&=(1-u_q)T_H.
\end{aligned}
\label{eq:temporal_alignment}
\end{equation}
Here, $T_q^+$ is the remaining duration and $t=t_h+vT_q^+$ is the physical time.
Predictions with $u_q\ge1$ have expired and are discarded.

We first fit the remaining prediction with eight B-spline control points
using regularized least squares. We translate all spatial controls by the
same offset so that the retained curve starts at the active-reference
position at $t_h$, without changing yaw or the remaining duration.
Let $\mathbf{C}_q^+\in\mathbb{R}^{8\times4}$ denote these reanchored controls.
We then adjust them by solving a constrained
least-squares problem. At the switch, the corrected trajectory must match
the current reference in position, velocity, and acceleration, as well as yaw,
yaw rate, and yaw acceleration. The objective keeps the corrected curve close
to the fitted prediction and penalizes changes in its third position derivative
and second yaw derivative to limit deformation.

The boundary conditions require derivatives with respect to physical time.
For the original prediction, $t=t_q+uT_H$ gives
\begin{equation}
    \frac{\mathrm{d}^{r}\hat{\boldsymbol{\tau}}_q}
         {\mathrm{d}t^{r}}
    =
    T_H^{-r}
    \frac{\mathrm{d}^{r}\hat{\boldsymbol{\tau}}_q}
         {\mathrm{d}u^{r}},
    \qquad r\in\{1,2\}.
    \label{eq:physical_derivatives}
\end{equation}
For the refitted remaining segment, the same conversion uses its coordinate
$v$ and duration $T_q^+$. The current reference and its derivatives are
evaluated at $t_h$.

For candidate controls
$\mathbf{C}$, let $\mathbf{p}_{\mathbf{C}}(s)$ and $\psi_{\mathbf{C}}(s)$ denote
position and yaw at local physical time $s\in[0,T_q^+]$, measured from $t_h$.
The corrected controls satisfy:
\begin{equation}
\begin{aligned}
    \mathbf{C}_q^{\mathrm{exec}}
    &=\arg\min_{\mathbf{C}\in\mathbb{R}^{8\times4}}
      \mathcal{L}_{\mathrm{fit}}
      \!\left(\mathbf{C},\mathbf{C}_q^{+}\right),\\[-1mm]
    \mathrm{s.t.}\quad
    \mathbf{p}_{\mathbf{C}}^{(r)}(0)
      &=\mathbf{p}_{\mathrm{act}}^{(r)}(t_h),
      \quad r=0,1,2,\\
    \psi_{\mathbf{C}}^{(r)}(0)
      &=\psi_{\mathrm{act}}^{(r)}(t_h),
      \quad r=0,1,2.
\end{aligned}
\label{eq:boundary_projection}
\end{equation}
In~\eqref{eq:boundary_projection}, superscript $(r)$ denotes differentiation
with respect to physical time.
These constraints provide $C^2$ continuity at the trajectory switch.

The boundary conditions specify how the new trajectory starts but leave
freedom in the rest of the curve. We select the corrected trajectory by
penalizing changes to the fitted remaining segment:
\begin{equation}
  \mathcal{L}_{\mathrm{fit}}
    =\mathcal{L}_{\mathrm{curve}}
     +\lambda\mathcal{L}_{\mathrm{deriv}}
     +\varepsilon\|\mathbf{C}-\mathbf{C}_q^+\|_F^2.
\label{eq:fusion_metric}
\end{equation}
Here, $\lambda$ weights the derivative correction term, and $\varepsilon$
controls control-point regularization.
The curve term measures weighted squared differences from the remaining
prediction, with greater weight on its later portion. The derivative term
penalizes changes in the third position derivative and second yaw derivative,
normalized relative to the curve term. These derivatives are taken with
respect to $v$. The final term penalizes changes to the control points.

At this projection stage, a fitted proposal that already satisfies the boundary
constraints remains unchanged because the objective penalizes only the correction.
The projection enforces boundary continuity while minimizing changes to the
retained trajectory.
Before activation, the executor checks speed, acceleration, and yaw-rate bounds using the spline's
derivative control points. A candidate that passes these checks becomes the
active reference at $t_h$, with duration $T_q^+$.
Rejected proposals and missed execution deadlines invoke a brake-to-hover
fallback from the active-reference boundary. After braking, a stationary
reference is maintained until the episode terminates.

\subsection{Execution-Aware GRPO}
\label{sec:execution-aware-ppo}
Accumulated closed-loop deviations shift subsequent observations away from
the demonstration distribution used for SFT. We therefore adapt the
policy to asynchronous execution using trajectory-level rewards and a
GRPO-style update~\cite{shao2024deepseekmath}.

For each task $i$, the pre-update policy $\pi_{\mathrm{old}}$ generates $K$
stochastic rollouts. We compute group-relative advantages from rollout returns
and use them in a token-level clipped objective. Pre-update log probabilities
are recorded for sampled action tokens associated with the executed trajectory.

\paragraph{Asynchronous trajectory reward}
We evaluate the complete executed trajectory against the expert demonstration
using both position and orientation. Dynamic time warping aligns the two
sequences to accommodate differences in execution speed. We use normalized
dynamic time warping (nDTW)~\cite{ilharco2019ndtw} as the trajectory-level
reward, with higher values indicating closer agreement.
Let $\boldsymbol{\rho}_i=(\rho_{i,1},\ldots,\rho_{i,M_i})$ and
$\boldsymbol{\chi}_{i,k}$ denote the corresponding feature sequences of the
demonstration and rollout $k$, respectively:
\begin{equation}
    \begin{aligned}
    \operatorname{DTW}(\boldsymbol{\rho}_i,\boldsymbol{\chi}_{i,k})
      &=\min_{W\in\mathcal{W}}
        \sum_{(m,n)\in W}\|\rho_{i,m}-\chi_{i,k,n}\|_2,\\
    L_i&=\sum_{m=2}^{M_i}\|\rho_{i,m}-\rho_{i,m-1}\|_2,\\
    \operatorname{nDTW}(\boldsymbol{\rho}_i,\boldsymbol{\chi}_{i,k})
      &=\exp\!\left(
        -\frac{\operatorname{DTW}(\boldsymbol{\rho}_i,
                                   \boldsymbol{\chi}_{i,k})}
        {L_i}\right),
    \end{aligned}
    \label{eq:ndtw_reward}
\end{equation}
Here, $\mathcal{W}$ contains sequence alignments that preserve temporal order,
and $L_i$ normalizes the discrepancy by the demonstration's path length in
the same feature space. At the end of rollout $k$, we assign
$R_{i,k}=\operatorname{nDTW}(\boldsymbol{\rho}_i,\boldsymbol{\chi}_{i,k})$.

\paragraph{Group-normalized advantages}
We screen rollout groups to avoid policy-gradient updates based solely on
failed or non-improving exploration. A group is retained only when at least
one valid rollout exceeds a reference return threshold. The validity criterion
and threshold are specified with the training settings.

For each retained group, the trajectory advantage is
\begin{equation}
    A_{i,k}
      =\frac{R_{i,k}-\bar R_i}
        {\max(\sigma_i,\eta)},
    \label{eq:signed_async_advantage}
\end{equation}
Here, $\bar R_i$ and $\sigma_i$ are the mean and standard deviation of the
group's returns, and $\eta>0$ sets a lower bound on the normalization
denominator. Both positive and negative advantages are retained. Each rollout's advantage is shared by its action tokens used in
the GRPO update.

\paragraph{Policy optimization}
The policy objective increases the probability of sampled actions with
positive advantages and decreases it for those with negative advantages.
For each action token, we compare its probability under the current and
pre-update policies, conditioned on the same observation and preceding
tokens, denoted by $h_{i,k,q,j}=(o_{i,k,q},\mathbf{y}_{i,k,q,<j})$.
Both probabilities are evaluated without sampling filters and at unit
temperature:
\begin{equation}
    r_{i,k,q,j}(\theta)
    =\frac{p_\theta(y_{i,k,q,j}\mid h_{i,k,q,j})}
    {p_{\mathrm{old}}(y_{i,k,q,j}\mid h_{i,k,q,j})}.
    \label{eq:token_ratio}
\end{equation}
We retain PPO-style probability-ratio clipping~\cite{schulman2017ppo} to constrain
policy updates, yielding the following GRPO-style objective:
\begin{equation}
\begin{aligned}
    \bar r_{i,k,q,j}(\theta)
      &=\operatorname{clip}\bigl(
        r_{i,k,q,j}(\theta),1-\epsilon_{-},1+\epsilon_{+}\bigr),\\
    \mathcal{L}_{\mathrm{clip}}(\theta)
      &=-\mathbb{E}_{i,k,q,j}\!\left[
        \min\!\left(r_{i,k,q,j}A_{i,k},
                     \bar r_{i,k,q,j}A_{i,k}\right)
      \right].
\end{aligned}
\label{eq:ppo_objective}
\end{equation}
The parameters $\epsilon_-$ and $\epsilon_+$ specify the lower and upper
clipping margins. The clipped policy loss is averaged over the action tokens used in
each prediction, then over predictions, rollouts, and tasks. Excluded groups
contribute zero to this loss.

To discourage excessive deviation from the pre-update policy, we add a
Kullback--Leibler (KL) penalty against a frozen copy, $\pi_{\mathrm{ref}}$.
The penalty is evaluated on observation and token contexts collected from
greedy reference rollouts, denoted by $\mathcal{H}_{\mathrm{ref}}$:
\begin{equation}
    \mathcal{L}
    =\mathcal{L}_{\mathrm{clip}}
    +\beta\mathbb{E}_{h\sim\mathcal{H}_{\mathrm{ref}}}
      \mathcal{D}_{\mathrm{KL}}\!\left(
        p_\theta(\cdot\mid h)\,\|\,p_{\mathrm{ref}}(\cdot\mid h)\right).
    \label{eq:full_ppo_objective}
\end{equation}
Here, $\beta\geq0$ controls the KL regularization strength. Only the policy's
LoRA adapters are updated; the backbone and reference-policy parameters
remain frozen.
\section{Experiments}
\label{sec:experiments}

\begin{figure*}[t]
  \centering
  \includegraphics[width=\textwidth]{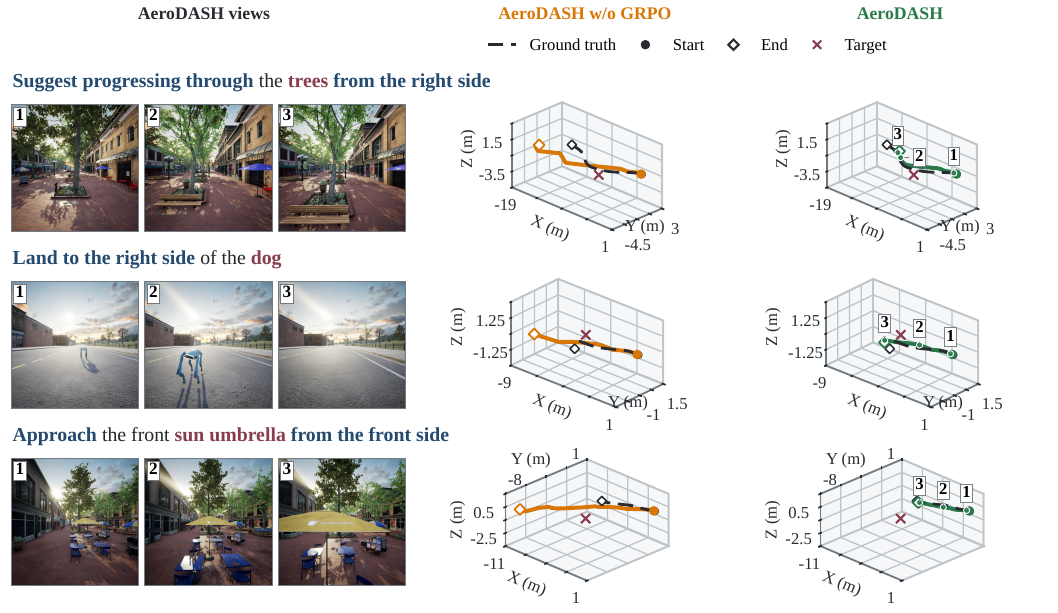}
  \caption{Language-conditioned flight before and after GRPO for Pass, Land, and Approach. Each row pairs three AeroDASH views with
  trajectories from AeroDASH w/o GRPO and AeroDASH. Dashed curves denote
  expert reference trajectories (ground truth), crosses mark referenced targets, and image indices identify
  the corresponding AeroDASH poses.}
  \label{fig:qualitative-trajectories}
\end{figure*}

\subsection{Experimental Setup}

\subsubsection{Dataset and Baselines}
We evaluate all methods on 273 tasks spanning ten instruction categories in
the UAV-Flow simulation test set~\cite{wang2025uavflow}. Initial conditions
and language instructions are identical across methods.

We compare AeroDASH with \emph{OpenVLA-UAV}~\cite{wang2025uavflow}, the
UAV-adapted OpenVLA baseline introduced in UAV-Flow. We additionally evaluate
a \emph{Waypoint Chunking} baseline that predicts seven waypoints per query.
Both this baseline and AeroDASH produce seven four-dimensional outputs
per query, represented as waypoints and B-spline control points, respectively. This comparison controls the number of predicted outputs
when evaluating the action representation and execution method.

OpenVLA-UAV uses its native sequential executor. The waypoint-chunk baseline,
AeroDASH w/o GRPO, and AeroDASH all use asynchronous execution.
The waypoint-chunk baseline tracks the predicted waypoints sequentially.
AeroDASH w/o GRPO uses the supervised
policy with the same B-spline representation and trajectory-fusion executor
as AeroDASH, isolating the contribution of GRPO post-training.

\subsubsection{Training and Implementation Details}
\label{sec:training-details}
Both the waypoint-chunk baseline and the AeroDASH B-spline policy are trained for
50,000 SFT steps. GRPO post-training is initialized from the resulting B-spline
policy and uses 2,730 tasks from the UAV-Flow-Sim training set. During GRPO,
only the LoRA adapters are updated; the backbone and reference policy remain
frozen. Training uses six NVIDIA RTX 4090 D GPUs.
We optimize the LoRA adapters with AdamW at a learning rate of $10^{-5}$,
using rank 32 and $\alpha=16$. GRPO uses clipping parameters $\epsilon_-=0.20$ and $\epsilon_+=0.28$,
and a KL penalty weight of 0.01. Each iteration
collects eight rollouts and one greedy reference rollout per task, with an
advantage normalization floor of $\eta=0.006$.
Rollouts use temperature-scaled sampling with top-$k$ and top-$p$ filtering.

During GRPO rollouts and navigation evaluation, each predicted trajectory covers
$T_H=1.4\,$s. We use a fixed inference delay of $0.6\,$s and switch trajectories
$0.7\,$s after each query. The controller receives reference updates at
$10\,$Hz while inference runs asynchronously.

A rollout terminates when the policy consistently predicts negligible motion
and the executed trajectory settles.

The greedy rollout return provides the reference $\mu_i^{\mathrm{ref}}$
for group selection. A group is retained for GRPO updates only if at least
one rollout completes normally and achieves $R_{i,k}>\mu_i^{\mathrm{ref}}+\eta$.
Excluded groups remain in the task average with zero GRPO contribution; the
KL term is computed separately as defined in~\eqref{eq:full_ppo_objective}.

\begin{table}[t]
  \caption{Performance on the UAV-Flow test set.}
  \label{tab:main-results}
  \centering
  \small
  \renewcommand{\arraystretch}{1.12}
  \setlength{\tabcolsep}{4.0pt}
  \begin{tabular*}{\columnwidth}{@{\extracolsep{\fill}}lcc@{}}
    \toprule
    Method & nDTW $\uparrow$ & SR (\%) $\uparrow$ \\
    \midrule
    OpenVLA-UAV~\cite{wang2025uavflow} & 0.2883 & 38.46 \\
    Waypoint Chunking & 0.3752 & 48.72 \\
    AeroDASH w/o GRPO & 0.3913 & 56.41 \\
    \midrule
    \textbf{AeroDASH} & \textbf{0.4495} & \textbf{76.56} \\
    \bottomrule
  \end{tabular*}
\end{table}

\subsubsection{Controlled Trajectory Execution}
\label{sec:execution-replay-details}

\begin{figure}[!t]
  \centering
  \includegraphics[width=\columnwidth]{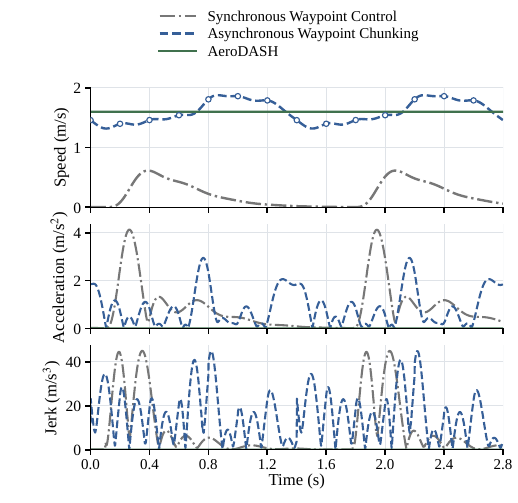}
  \caption{Comparison of three execution strategies in a straight-flight simulation
  with a nominal reference speed of $1.6\,$m/s. Panels show speed, acceleration
  magnitude, and jerk magnitude from top to bottom. Time is relative to the common $5.8$--$8.6\,$s
  excerpt; circles mark $0.2\,$s asynchronous waypoint updates.}
  \label{fig:execution-profiles}
  \par\vspace{8pt}
  \begingroup
  \makeatletter
  \def\@captype{table}
  \makeatother
  \caption{Execution-Strategy Comparison.}
  \label{tab:execution-comparison}
  \centering
  \small
  \renewcommand{\arraystretch}{1.20}
  \setlength{\tabcolsep}{2pt}
  \begin{tabular*}{\columnwidth}{@{\extracolsep{\fill}}lrrrr@{}}
    \toprule
    Method & $\bar v$ & $\sigma_v\downarrow$ & $a_{95}\downarrow$ & $j_{95}\downarrow$ \\
    & (m/s) & (m/s) & (m/s$^2$) & (m/s$^3$) \\
    \midrule
    Synchronous Waypoint Control & 0.148 & 0.1525 & 2.555 & 31.169 \\
    Waypoint Chunking & 1.198 & 0.1399 & 1.786 & 28.641 \\
    \textbf{AeroDASH} & 1.200 & \textbf{0.0009} & \textbf{0.184} & \textbf{0.212} \\
    \bottomrule
  \end{tabular*}
  \endgroup
\end{figure}

Figure~\ref{fig:execution-profiles} and Table~\ref{tab:execution-comparison}
compare three execution strategies across 40 conditions: five path families,
four geometric instances, and two nominal speeds ($0.8$ and $1.6\,$m/s).
The families comprise straight paths, circular arcs, S-bends, ascent/descent,
and rounded turns.

All three strategies use shared position/velocity feedback gains and a
rigid-body/motor model. Reference commands are published at $50\,$Hz,
feedback runs at $100\,$Hz, and dynamics are integrated at $1\,$kHz.
The position gain is 2, with velocity proportional/integral gains of 6 and 1.
Waypoint Chunking updates targets every $0.2\,$s without an arrival or settling
requirement; targets have zero velocity/acceleration feed-forward.
The asynchronous pair uses identical proposals with a $1.4\,$s horizon, $0.6\,$s inference
delay, and $0.7\,$s reference activation cadence; waypoint chunks activate
at the next target-update tick after becoming eligible.

In the execution comparison, AeroDASH removes the expired prediction prefix
and aligns the retained segment with the active-reference position at the switch.
It optimizes the retained trajectory for smoothness while maintaining $C^2$
continuity at the switching boundary.
The controller receives analytic position, velocity, and acceleration references.

Synchronous waypoint control uses the same controller and vehicle model.
It holds the target during $0.6\,$s inference,
triggered by position error $\leq0.05\,$m and speed $\leq0.05\,$m/s.
Each query generates a $0.2\,$s lookahead displacement from the same path rule,
anchored at the measured query position.

\newpage

\begin{figure}[!t]
  \centering
  \includegraphics[width=\columnwidth]{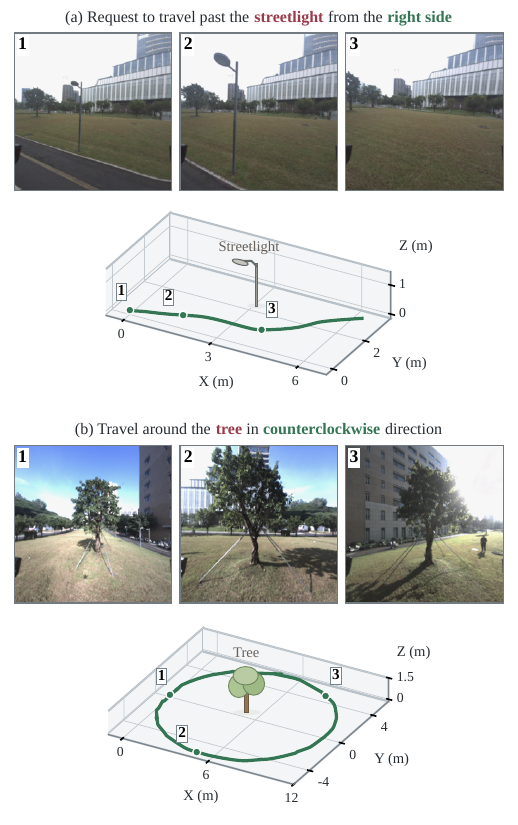}
  \caption{Onboard views and trajectory visualizations for streetlight passing
  and tree circling. The trajectories are reconstructed from measured UAV positions
  using cubic B-spline interpolation. Numbered markers link
  the curves to onboard images; object symbols indicate the locations of the streetlight and tree.}
  \label{fig:real-world-flight}
\end{figure}
\subsubsection{Onboard Deployment}
The UAV carries a Jetson Orin NX with 16\,GB memory, a forward-facing RGB
camera, and a PX4 flight controller. Super-LIO~\cite{wang2026superlio}
estimates the UAV state from a Livox MID-360 LiDAR and its IMU.
Images are resized to $224\times224$ for policy inference. Through ROS~2
and MAVROS, the system supplies odometry and continuous trajectory references
to PX4 at $50\,$Hz, independently of policy inference.

The onboard policy directly predicts B-spline control
points. We use TensorRT for inference and merge
the LoRA adapters before model export.
Activation-aware weight quantization (AWQ)~\cite{lin2024awq}
reduces the language model to 4-bit weights and 16-bit activations (W4A16),
while the visual encoder and multimodal projector use mixed FP16/FP32
precision. The deployed policy has a mean inference latency of $672\,$ms
on the Orin NX.

\subsubsection{Evaluation Metrics}
We report success rate (SR) and nDTW~\cite{ilharco2019ndtw}.
SR is the percentage of flights judged by manual review to satisfy the
language instruction. We compute nDTW using the UAV-Flow evaluator for all
methods, following the same feature representation and normalization used
for the training reward in~\eqref{eq:ndtw_reward}.

We compute execution metrics from vehicle-state histories generated by the
dynamics simulation. Jerk is calculated by applying central differences directly
to the raw acceleration samples at $1\,$ms intervals. For each condition, we calculate mean
speed $\bar v$, speed standard deviation $\sigma_v$, and the 95th-percentile
acceleration and jerk magnitudes $a_{95}$ and $j_{95}$ over $[2.8,9.8)\,$s.
Table~\ref{tab:execution-comparison} reports the equal-weight average of each
metric across all 40 conditions.

\subsection{Quantitative Comparison}

Table~\ref{tab:main-results} reports the closed-loop comparison on the
UAV-Flow test set. AeroDASH achieves $76.56\%$ SR and $0.4495$ nDTW,
compared with $38.46\%$ and $0.2883$ for OpenVLA-UAV, respectively.
The SR improvement is $38.1$ percentage points.

With both policies predicting seven four-dimensional outputs and using
asynchronous execution, AeroDASH w/o GRPO achieves $56.41\%$ SR and $0.3913$
nDTW, compared with $48.72\%$ and $0.3752$ for the waypoint-chunk baseline.
This corresponds to gains of $7.69$ percentage points in SR and $0.0161$ in nDTW.

GRPO post-training further increases SR from
$56.41\%$ to $76.56\%$ and nDTW from $0.3913$ to $0.4495$, yielding
absolute gains of $20.1$ percentage points and $0.0582$, respectively.

\subsection{Qualitative Case Analysis}

\subsubsection{Language-Conditioned Task Behavior}
In Fig.~\ref{fig:qualitative-trajectories}, AeroDASH follows the expert reference
path more closely in Pass and exhibits less target-directed overshoot in
Approach than AeroDASH w/o GRPO. In Land, AeroDASH achieves closer path
agreement with the expert reference.

\subsubsection{Real-World Flight}
Figure~\ref{fig:real-world-flight} presents onboard image sequences and corresponding
three-dimensional trajectory visualizations for streetlight passing and
counterclockwise tree circling.

\subsection{Execution-Strategy Comparison}

Figure~\ref{fig:execution-profiles} compares vehicle motion under inference
latency with shared controller and vehicle-model settings. Synchronous waypoint control
produces stop-and-wait motion, averaging $0.148\,$m/s (Table~\ref{tab:execution-comparison}).
Asynchronous waypoint-chunk control maintains
motion but exhibits larger speed fluctuations and higher acceleration and jerk peaks than AeroDASH.

For the asynchronous pair, Table~\ref{tab:execution-comparison} shows that
AeroDASH reduces $a_{95}$, $j_{95}$, and $\sigma_v$ by $89.7\%$,
$99.3\%$, and $99.3\%$, respectively, relative to Waypoint Chunking.
All three metrics improve in every condition; mean speeds remain comparable
at $1.200$ and $1.198\,$m/s, respectively.

\section{Conclusion}
\label{sec:conclusion}

We presented AeroDASH, an asynchronous VLA framework for continuous
language-conditioned UAV flight. Trajectory fusion aligns delayed B-spline
predictions with the active reference while maintaining derivative continuity
between successive updates. Execution-aware GRPO further adapts the policy to
closed-loop deviations through trajectory-level feedback. UAV-Flow experiments
demonstrate improved task success. In execution simulations, AeroDASH maintains
continuous motion with lower acceleration and jerk than waypoint-based execution.
Onboard flights further demonstrate language-conditioned control with policy
inference and trajectory execution running concurrently.

\end{document}